\documentclass[preprint,12pt]{elsarticle}

\usepackage{amssymb}
\usepackage{amsmath}
\usepackage{multirow}

\usepackage{hyperref}
\hypersetup{
	colorlinks=true,
	linkcolor=cyan,
	filecolor=blue,      
	urlcolor=red,
	citecolor=green,
}

\journal{arXiv}

\begin{document}

\begin{frontmatter}



\title{Faster Diffusion Action Segmentation}


\address[add1]{Academy for Engineering and Technology, Fudan University, Shanghai, 200433, China}
\address[add2]{Engineering Research Center of AI and Robotics, Ministry of Education, Shanghai, 200433, China}
\fntext[equal]{Equal contribution.}
\cortext[mycorrespondingauthor]{Corresponding author.}

\author[add1,add2]{Shuaibing Wang\fnref{equal}}
\author[add1,add2]{Shunli Wang\fnref{equal}}
\author[add1,add2]{Mingcheng Li}
\author[add1,add2]{Dingkang Yang}
\author[add1,add2]{Haopeng Kuang}
\author[add1,add2]{Ziyun Qian}
\author[add1,add2]{Lihua Zhang\corref{mycorrespondingauthor}}
\ead{sbwang21@m.fudan.edu.cn, lihuazhang@fudan.edu.cn}

\begin{abstract}

Temporal Action Segmentation (TAS) is an essential task in video analysis, aiming to segment and classify continuous frames into distinct action segments. However, the ambiguous boundaries between actions pose a significant challenge for high-precision segmentation. Recent advances in diffusion models have demonstrated substantial success in TAS tasks due to their stable training process and high-quality generation capabilities. 
However, the heavy sampling steps required by diffusion models pose a substantial computational burden, limiting their practicality in real-time applications. Additionally, most related works utilize Transformer-based encoder architectures. Although these architectures excel at capturing long-range dependencies, they incur high computational costs and face feature-smoothing issues when processing long video sequences.
To address these challenges, we propose EffiDiffAct, an efficient and high-performance TAS algorithm. 
Specifically, we develop a lightweight temporal feature encoder that reduces computational overhead and mitigates the rank collapse phenomenon associated with traditional self-attention mechanisms. Furthermore, we introduce an adaptive skip strategy that allows for dynamic adjustment of timestep lengths based on computed similarity metrics during inference, thereby further enhancing computational efficiency.
Comprehensive experiments on the 50Salads, Breakfast, and GTEA datasets demonstrated the effectiveness of the proposed algorithm.

\end{abstract}

    





\end{frontmatter}


\section{Introduction}

\begin{figure}[htbp]
    \centering
    \includegraphics[width=0.9\textwidth]{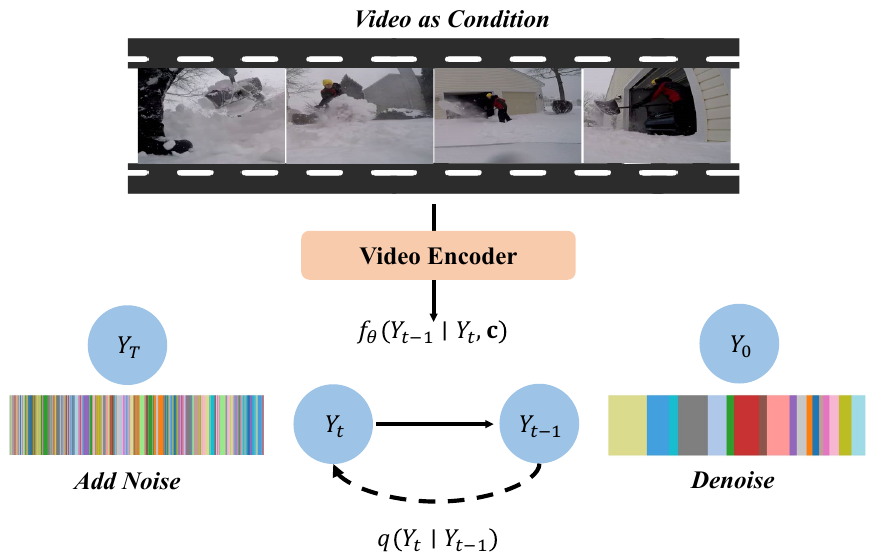}
    
    \caption{The application process of diffusion models in TAS tasks.}
    \label{fig1}
\end{figure}

Temporal Action Segmentation (TAS) is a crucial task in the field of video analysis, with the core objective of segmenting and classifying continuous frames in a video stream into different action segments. However, the ambiguous boundaries between actions make high-precision segmentation a challenging problem. In recent years, diffusion models have achieved significant success in the generative model domain due to their stable training process and high-quality generation capabilities. These models inject noise into the data progressively in the forward process to disrupt its structure and then gradually remove the noise in the reverse process, ultimately aiming to reconstruct new data samples from a highly random noise state. This iterative denoising process naturally aligns with the iterative refinement paradigm of action segmentation and provides a new perspective for video analysis. Liu \textit{et al.} \cite{liu2023diffusion} indicates that diffusion models can significantly improve the accuracy of temporal action segmentation. Figure \ref{fig1} illustrates the specific application process of diffusion models in the task of temporal action segmentation, where the video frame sequence serves as a conditional input, generating action labels from a pure noise action sequence.
However, when applying diffusion models to real-world scenarios, their burdensome sampling steps become a significant limiting factor. Specifically, the sampling process of diffusion models typically involves multiple iterative steps to recover the target data from the noise distribution gradually. Although this approach achieves significant results in generation quality, its inherent iterative process requires extensive forward and reverse computations, resulting in slow processing speeds. Consequently, in applications requiring real-time analysis, computational delays become a major bottleneck.

Furthermore, most related research employs Transformer-based encoder architectures. Despite their inherent advantage in capturing long-range dependencies, still face high computational costs and feature smoothing issues when dealing with long video data.
Unlike other fields, such as machine translation, the features of input video segments in temporal action segmentation tasks are often very similar. When integrating wide-range contextual information, Transformer models may excessively weaken the distinctiveness of local features, thereby reducing the model’s sensitivity to subtle action changes in videos.

For temporal action segmentation tasks, the most effective temporal context modeling should retain key information and highlight the most distinctive video features. Additionally, Shi \textit{et al.} \cite{shi2023tridet} pointed out that removing the self-attention mechanism does not significantly degrade model performance, suggesting that simply relying on Transformer-encoded features does not significantly improve action recognition accuracy. This study delved into the phenomenon known as the "rank collapse" issue, revealing that the self-attention mechanism might cause the input feature matrix to converge to rank 1 quickly. In short, the self-attention mechanism may diminish the distinctiveness of features at each time step, causing the input sequence features to exhibit increasingly high similarity. In temporal action segmentation tasks, significant distinctions between features are crucial for understanding and recognizing different action segments. This feature smoothing is clearly detrimental to the accurate identification and analysis of actions.

To address the above issues, this study proposes the EffiDiffAct algorithm, which builds upon DiffAct \cite{liu2023diffusion}. EffiDiffAct achieves this with a lightweight encoder and an adaptive skip strategy. These innovations aim to optimize the computational efficiency of the diffusion model in temporal action segmentation tasks.

The main contributions are summarized as follows:

\begin{itemize}
\item We propose a lightweight feature encoder that not only significantly reduces the computational burden of the model when processing long videos but also effectively mitigates the issue of excessive temporal smoothing in features caused by traditional self-attention mechanisms, known as the "rank collapse" phenomenon. This module enables the model to distinguish features at different moments in the video while reducing computational resource consumption, thereby improving the accuracy of action segmentation.

\item We propose an adaptive skip strategy that allows the model to dynamically adjust the skipping length based on the similarity metrics computed during inference, in contrast to the fixed skipping strategy used in traditional diffusion models. This strategy optimizes the selection of time steps to further enhance the model's inference speed without sacrificing prediction quality.

\item Qualitative and quantitative experiments conducted on multiple public datasets demonstrate the efficiency and effectiveness of our proposed algorithm.
\end{itemize}

\section{Related Work}
Temporal action segmentation aims to automatically identify and delineate different action segments from a video sequence. Assume the video sequence can be represented as a collection of frames \( V = \{ v_1, v_2, \ldots, v_T \} \), where \( T \) is the total number of frames in the video, and \( v_t \) represents the feature vector of the \( t \)-th frame. The goal of temporal action segmentation is to partition the video sequence \( V \) into \( N \) action segments, each composed of a series of consecutive frames corresponding to a specific action category. Define a function \( f \) that maps the video sequence \( V \) to an action label sequence \( S = \{ s_1, s_2, \ldots, s_T \} \), i.e., \( S = f(V) \), where \( s_t \) denotes the action label of the \( t \)-th frame.

\textbf{Traditional methods.} In the early research on action segmentation, sliding window methods \cite{rohrbach2012database, tanberk2023supervised} were widely used for detecting action segments, filtering redundant segments through Non-Maximum Suppression (NMS). The primary limitation of these methods lies in the setting of the window size, making it difficult to adapt to variations in action duration and capture long-term dependencies of action sequences. Other studies have employed models such as Hidden Markov Model (HMM) \cite{tang2012learning, kuehne2016end}, Conditional Random Field (CRF) \cite{pirsiavash2014parsing, shi2011human, tao2013surgical}, and Recurrent Neural Network (RNN) \cite{donahue2015long, yue2015beyond, singh2016multi, vinyals2015show} to model the temporal relationships of actions. These approaches aim for a more effective representation of inter-frame associations to achieve precise classification of each frame in the video. Although HMM and CRF show some effectiveness in modeling short-term dependencies, they struggle with long-term dependencies. While RNNs can handle longer sequences, they are affected by the vanishing or exploding gradient problem in processing long video data. Additionally, the structure of RNNs limits their ability for parallel computation.

\textbf{TCN-based methods.} In recent years, Temporal Convolutional Networks (TCNs) \cite{lea2017temporal} and Transformer models \cite{vaswani2017attention} have been applied to the field of temporal action segmentation, demonstrating impressive performance in handling video action sequences. MS-TCN \cite{farha2019ms} decomposes the temporal action segmentation task into multiple stages, with each stage using an independent TCN to capture action features at different temporal scales, progressively refining the recognition and classification of actions. However, TCNs still show certain limitations in handling global contextual information and understanding the global dependencies of the entire action sequence.

\textbf{Transformer-based methods.} ASFormer \cite{yi2021asformer} is the first temporal action segmentation model based on the Transformer architecture. It leverages the powerful self-attention mechanism of the Transformer to capture the global dependencies of actions within the video, achieving precise temporal segmentation throughout the entire action sequence. Compared to traditional CNN or RNN-based methods, Transformers can more effectively handle long-term dependencies and establish complex relationships across the entire action sequence. However, Transformers have higher computational complexity when processing long videos, and the "rank collapse" problem \cite{shi2023tridet} that arises in modeling temporal features further limits their application in certain scenarios.

\textbf{Diffusion-based methods.} Meanwhile, diffusion models \cite{ho2020denoising, song2020denoising} have achieved remarkable results in image generation, natural language generation, audio generation, and other fields. They have gradually been applied to some image understanding tasks in computer vision, such as object detection \cite{chen2023diffusiondet} and image segmentation \cite{amit2021segdiff}. Diffusion models aim to approximate the data distribution \( q(x_0) \) with the model distribution \( p_\theta(x_0) \). The generation process involves two stages: forward diffusion and reverse denoising. The forward process transforms real data \( x_0 \sim q(x_0) \) into a series of noisy data \(\{x_1, x_2, \ldots, x_s\} \). The reverse process gradually removes noise from \( x_s \sim \mathcal{N}(\mathbf{0}, \mathbf{I}) \), iterating through \(\{x_{s-1}, x_{s-2}, \ldots, x_1\} \) to finally obtain \( x_0 \sim p_\theta(x_0) \). This involves first "destroying" the data and then "recovering" it. The iterative denoising process of diffusion models is extremely similar to the progressive refinement of action predictions in temporal action segmentation algorithms, making diffusion models particularly suitable for temporal action segmentation tasks.

\section{EffiDiffAct}

\begin{figure}[h]
    \centering
    \includegraphics[width=\textwidth]{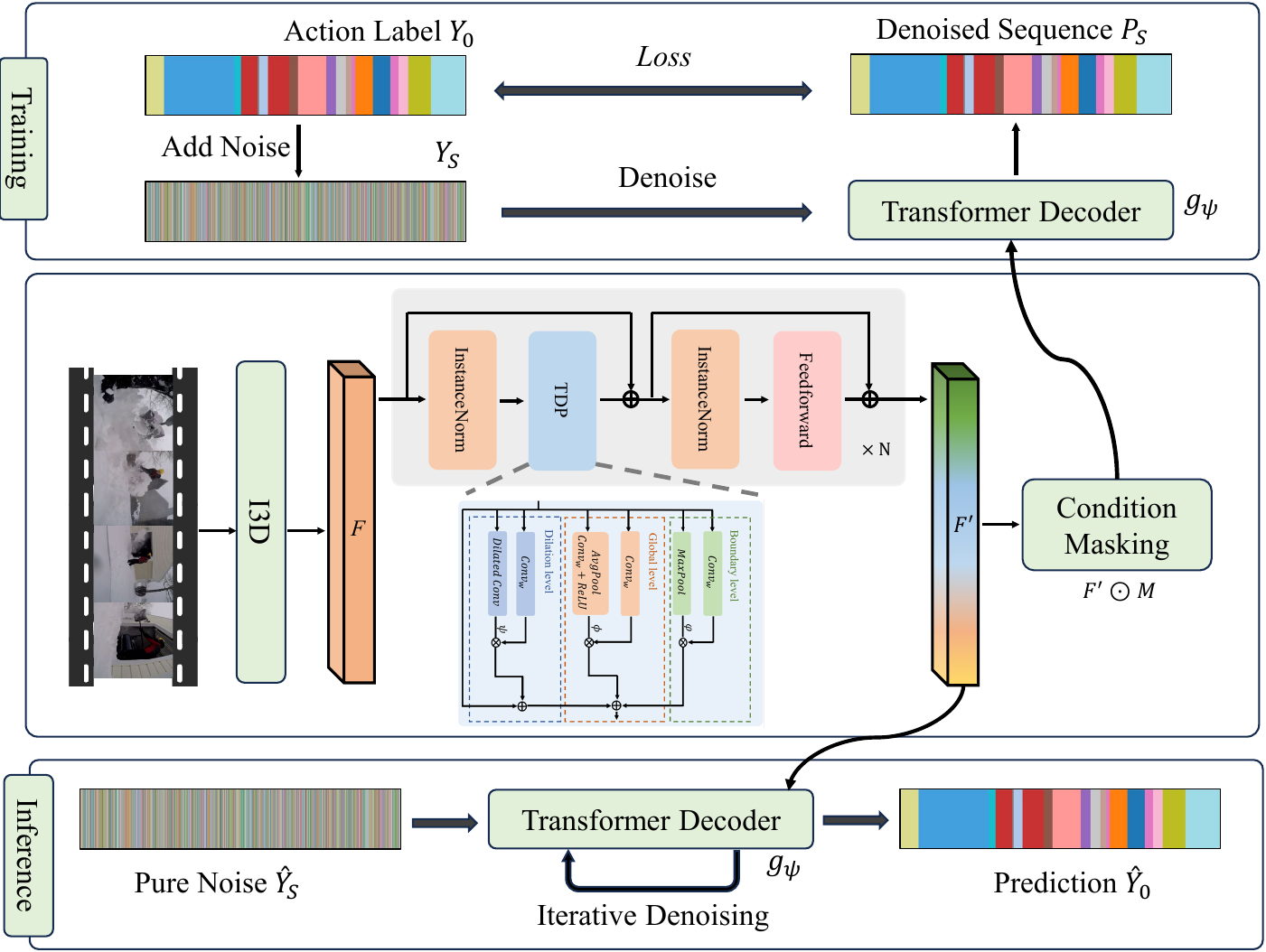}
    
    \caption{The overall framework of the EffiDiffAct algorithm. Given video features as conditional information, the model learns the mapping between noise sequences and action labels during the training phase, and during the inference phase, it restores action labels  $\hat{Y}_{0}$ from noise sequences $\hat{Y}_{S}$.}
    \label{fig3}
\end{figure} 

\vspace{5mm}
The overall framework of the EffiDiffAct is shown in Figure \ref{fig3}. The EffiDiffAct algorithm takes a video containing multiple action segments as input and utilizes a pre-trained action classification network (such as I3D \cite{carreira2017quo} or SlowFast \cite{feichtenhofer2019slowfast}) to extract the video's feature representation \( F \in \mathbb{R}
^{L \times 2048}\). Subsequently, the features \( F \) are passed through a TDP (Temporal Dilation Perception) feature encoder \( h_\phi \), which consists of multiple TDP layers. This encoder not only maximizes the information of the videos but also more effectively handles long-term temporal dependencies. During the training phase, Gaussian noise is added to the video action labels \( Y_0 \), generating a noisy sequence \( Y_s \). This noisy sequence \( Y_s \), along with the features processed by the conditional mask \cite{liu2023diffusion}, is input into the decoder \( g_\psi \). The task of the decoder \( g_\psi \) is to progressively restore the noise-free action label sequence, outputting the denoised sequence \( P_s \). During the inference phase, the algorithm starts with a purely noisy sequence \( \hat{Y}_S \) and iteratively denoises it through the decoder \( g_\psi \), ultimately obtaining the predicted action sequence \( \hat{Y}_{0} \).

\textbf{Training Phase.} The diffusion process starts with the original action label sequence \( Y_0 \). In each training iteration, a diffusion step \( s \) is randomly selected from the set \(\{1,2,\ldots,S\}\), and noise is added to \( Y_0 \) according to the cumulative noise value \( \overline{\alpha}_s \), resulting in the corrupted sequence \( Y_s \in \{0,1\}^{L\times C} \):
\begin{equation}
    Y_s = \sqrt{\overline{\alpha}_s} Y_0 + \epsilon \sqrt{1-\overline{\alpha}_s}, \quad \epsilon \sim N(0,I). \label{equ1}
\end{equation}

It is noteworthy that the encoding phase of the model relies on the TDP encoder proposed in this paper, while the decoding phase still employs the Transformer decoder. This is because the encoding phase only needs to process video features, whereas the decoding phase requires interaction between the video features and the noise sequence, which the TDP encoder does not support. Therefore, this study uses the Transformer decoder \( g_\psi \) to denoise the sequence and recover the action label sequence:

\begin{equation}
P_s = g_\psi\left(Y_s, s, h_\phi(F) \odot M\right), \label{equ2}
\end{equation}
where \( P_s \in [0,1]^{L \times C} \) represents the action probability distribution for each frame of the video, where \( L \) is the length of the video sequence and \( C \) is the number of action categories. The inputs to the decoder \( g_\psi \) include the noise sequence \( Y_s \), the current step \( s \) of the diffusion process, and the encoded video features \( h_\phi(F) \). The step \( s \) provides the decoder with information about the stage of the denoising process, and \( h_\phi(F) \) provides the contextual information needed by the decoder to recover the action probability distribution more accurately.

Furthermore, we employ a conditional masking strategy \cite{liu2023diffusion}, where the video features \( h_\phi(F) \) are element-wise multiplied by a mask \( M \) (denoted as \(\odot\), the Hadamard product) to control the conditional information input to the model. The conditional masking strategy leverages three priors of human actions: positional prior, boundary prior, and relational prior, by introducing different types of masks. By randomly applying an all-ones mask, an all-zeros mask, a boundary mask, and a relational mask, this strategy allows the model to switch between fully conditional information, no conditional information, and partial conditional information flexibly. This strengthens the model's understanding of action sequences during training and enhances its ability to capture action distributions from multiple dimensions.

Based on Equation (\ref{equ2}), the decoder \( g_\psi \) can effectively recover the action probability distribution \( P_s \) from noisy data, accurately predicting the action category for each frame in the video and achieving high-precision action segmentation. To optimize the model, this paper utilizes a combination of three loss functions to measure the difference between \( P_s \) and \( Y_0 \): the cross-entropy loss \( \mathcal{L}_s^{ce} \), the temporal smoothness loss \( \mathcal{L}_s^{smo} \), and the boundary alignment loss \( \mathcal{L}_s^{bd} \). Therefore, the total loss function \( \mathcal{L}_{\mathrm{total}} \) considers the model's performance in predicting action categories, maintaining the smoothness of the time series, and accurately identifying action boundaries. It is defined as follows:
\begin{equation}
    \mathcal{L}_{\mathrm{total}} = \mathcal{L}_s^{ce} + \mathcal{L}_s^{smo} + \mathcal{L}_s^{bd}. \label{equ3}
\end{equation}

The cross-entropy loss \( \mathcal{L}_s^{ce} \) focuses on measuring the difference between the predicted probability distribution and the actual labels:
\begin{equation}
\mathcal{L}_s^{ce} = \frac{1}{LC} \sum_{i=1}^{L} \sum_{c=1}^{C} -Y_{0,i,c} \log P_{s,i,c}, \label{equ4}
\end{equation}
where \( L \) represents the number of video frames, \( C \) represents the number of action categories, \( Y_{0,i,c} \) is the true action label (one-hot encoded), and \( P_{s,i,c} \) is the model's predicted probability for frame \( i \) and category \( c \) at step \( s \).

The temporal smoothness loss \( \mathcal{L}_s^{smo} \) encourages consistency between action predictions of adjacent frames. It helps ensure the model's output is continuous and smooth over time, reducing abrupt changes in predictions and ensuring natural transitions between actions. The formula is as follows:

\begin{equation}
\mathcal{L}_s^{smo} = \frac{1}{(L-1)C} \sum_{i=1}^{L-1} \sum_{c=1}^{C} \left(\log P_{s,i,c} - \log P_{s,i+1,c}\right)^2. \label{equ5}
\end{equation}

The boundary alignment loss \( \mathcal{L}_s^{bd} \) aims to improve the model's accuracy in identifying action boundaries, i.e., the start and end points of actions. To achieve this, action boundary probabilities need to be derived from the denoised sequence \( P_s \) and the true action sequence \( Y_0 \). First, a true action boundary sequence \( B \in \{0,1\}^{L-1} \) is derived from \( Y_0 \), where \( B_i = 1 \) if \( Y_{0,i} \neq Y_{0,i+1} \). Since action transitions are usually gradual, a Gaussian filter is applied to smooth this sequence, yielding a softened version \( \overline{B} = \lambda(B) \). The boundary probability of the denoised sequence \( P_s \) is obtained by calculating the dot product of the action probabilities of adjacent frames in \( P_s \), i.e., \( 1 - P_{s,i} \cdot P_{s,i+1} \). Finally, the two boundaries are aligned using binary cross-entropy loss:

\begin{equation}
\mathcal{L}_s^{bd} = \frac{1}{L-1} \sum_{i=1}^{L-1} \left( -\overline{B}_i \log (1 - P_{s,i} \cdot P_{s,i+1}) - (1 - \overline{B}_i) \log (P_{s,i} \cdot P_{s,i+1}) \right). \label{equ6}
\end{equation}

\textbf{Inference Phase.} The core of this phase relies on the mapping relationship learned by the model during the training phase, from noise to clear sequences. The denoising process starts with a pure noise sequence \( \hat{Y}_S \), which is denoised by the decoder \( g_\psi \) to obtain \( \hat{Y}_{S-1} \). The denoised sequence \( \hat{Y}_{S-1} \) is then input back into the decoder, undergoing a series of iterative refinement steps, eventually recovering the action sequence \( \hat{Y}_0 \). Each iteration step follows the formula:

\begin{equation}
\hat{Y}_{s-1} = \sqrt{\overline{\alpha}_{s-1}} P_s + \sqrt{1 - \overline{\alpha}_{s-1} - \sigma_s^2} \Bigg(\frac{\hat{Y}_s - \sqrt{\overline{\alpha}_s} P_s}{\sqrt{1 - \overline{\alpha}_s} } \Bigg) + \sigma_s \epsilon, \label{equ7}
\end{equation}
where \( \epsilon \) is sampled from a normal distribution \( \mathcal{N}(0, I) \).

\subsection{Temporal Dilation Perception Encoder}
\begin{figure}[h]
    \centering
    \includegraphics[width=0.9\textwidth]{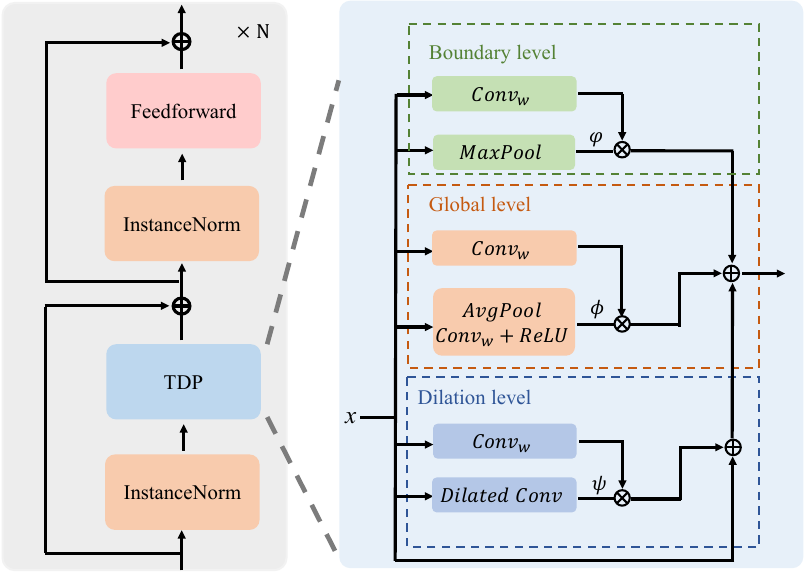}
    
    \caption{The overall framework of the TDP encoder, which maximizes video information through three levels: boundary level, global level, and dilation level.}
    \label{fig4}
\end{figure} 
To improve the computational efficiency of the model and enhance the temporal contextual representation of video features, this paper proposes the Temporal Dilation Perception (TDP) encoder, as shown in Figure \ref{fig4}. This module is based on the Transformer architecture but introduces TDP layers to replace the traditional self-attention mechanism, addressing the high computational cost and potential over-smoothing of features when processing long video sequences with Transformer encoders. The TDP encoder consists of multiple stacked TDP layers, each beginning with Instance Normalization (IN), which helps reduce internal covariate shift during training. For temporal feature modeling, the TDP layer employs three strategies: boundary level, global level, and dilation level.

\textbf{Boundary level.} The boundary level uses max pooling to extract the most prominent features in the sequence, highlighting key action boundaries. This design allows the boundary level to maintain sensitivity to critical instantaneous actions while preserving essential temporal information, effectively preventing the over-smoothing of features.

\textbf{Global level.} The primary function of the global level is to aggregate statistical information over the entire time sequence, providing a comprehensive view for the model. This is achieved by performing global average pooling along the temporal dimension, effectively reducing noise interference and capturing the video content's overall context. Subsequently, a 1D depthwise separable convolution layer \cite{chollet2017xception} is used to process these global features further, and a ReLU activation function is introduced to add non-linearity. This level is crucial for understanding long-term action trends in the video, offering a macroscopic understanding of the entire sequence and helping the model grasp global content while analyzing local details.

\textbf{Dilation level.} Considering that the temporal action segmentation task requires a deep understanding of long-term temporal dependencies, this paper further proposes the dilation level to achieve this goal. Traditional methods often rely on the self-attention mechanism of Transformers to model temporal dependencies, but the self-attention mechanism faces high computational costs when processing long videos. In contrast, the dilation level can effectively capture long-range temporal dependencies while improving computational efficiency. Its core idea involves using dilated convolutions \cite{yu2015multi}, which expand the spacing within the convolutional kernel to cover a longer temporal span without adding extra parameters compared to traditional convolutions, thereby capturing long-term temporal dependencies. 

Additionally, by using different dilation rates in various TDP layers, the model can capture multi-scale temporal features, enhancing the overall representation of the temporal sequence. This multi-scale temporal feature capture method allows the model to learn features at different time scales at different levels, thus providing a more comprehensive understanding and analysis of temporal features. The TDP layer can be expressed with the following formula:
\begin{equation}
f_{\text {TDP }}=\varphi(x) \operatorname{Conv}_w(x)+\phi(x) \operatorname{Conv}_w(x)+\psi(x) \operatorname{Conv}_w+x \label{equ8},
\end{equation}
where ${\rm Conv}_w$ denotes a one-dimensional depthwise separable convolution applied along the time dimension with a window size of $w$ \cite{chollet2017xception}, which effectively captures local temporal features. $\varphi(x)$, $\phi(x)$, and $\psi(x)$ represent the core operations of the TDP's three strategies by respectively multiplying them with ${\rm Conv}_w(x)$, the model can process and refine the input $x$ in different ways at each level. This allows each level to selectively process the input data, capturing different temporal scale information. Combining the three strategies enables the model to integrate global context information while dealing with local details. The computation processes of $\varphi(x)$, $\phi(x)$, and $\psi(x)$ are represented as follows:

\begin{equation}
\begin{aligned}
&\varphi(x)=\operatorname{MaxPool}(x), \\
&\phi(x)=\operatorname{ReLU}\big(\operatorname{Conv}_w\left(\operatorname{AvgPool}(x)\right)\big), \\
&\psi(x)=\operatorname{DilatedConv}(x). \label{equ9}
\end{aligned}
\end{equation}

Combining the above three strategies enables the model to capture and understand long-term patterns and complex actions in video sequences at different levels. By stacking these hierarchically modeled features and integrating them with the original input using residual connections, the model not only alleviates the vanishing gradient problem in deep networks but also maintains sensitivity to the original input features. The stacked features are further processed through instance normalization to ensure consistency in feature distribution before entering the next module. Subsequently, a Feedforward Neural Network (FNN) further processes the temporal features, enhancing their representation capabilities. The features output by the TDP encoder, after being processed by conditional masking, serve as conditional features, providing the necessary contextual information for the decoding stage to predict action sequences accurately.

\subsection{Adaptive Skip Strategy}

\begin{figure}[h]
    \centering
    \includegraphics[width=\textwidth]{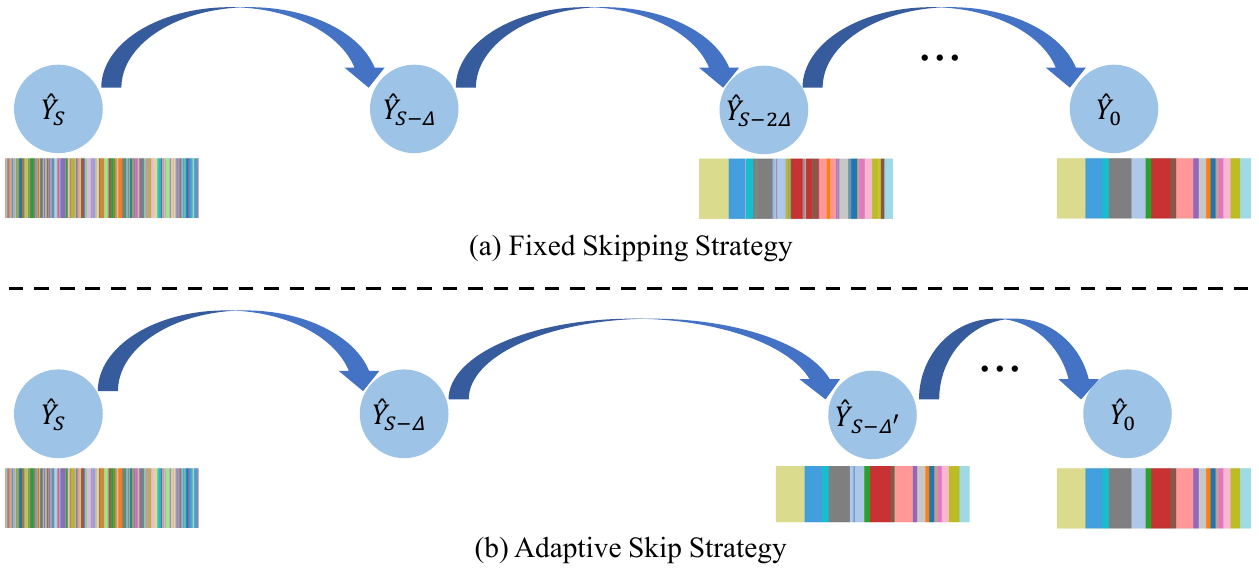}
    
    \caption{Comparison diagram of fixed skip strategy and adaptive skip strategy.}
    \label{fig5}
\end{figure}

DiffAct \cite{liu2023diffusion} adopts the Denoising Diffusion Implicit Model (DDIM) \cite{song2020denoising} to accelerate the inference process. DDIM introduces the concept of nonlinear time steps, allowing multiple time steps to be skipped during the generation process. Specifically, this strategy allows the model to jump directly from time step \( S \) to \( S - \Delta \), forming a nonlinear time sequence \(\{\hat{Y}_S, \hat{Y}_{S-\Delta}, \ldots, \hat{Y}_0\}\), thereby accelerating the generation process. Although this fixed skip strategy improves inference efficiency to some extent, it lacks sufficient flexibility, leading to a waste of computational resources.

To overcome the limitations of the fixed skip strategy, we propose an adaptive skip strategy aimed at further improving sampling efficiency while ensuring prediction quality, as illustrated in Figure \ref{fig5}. Specifically, this strategy dynamically adjusts the time step \(\Delta\) based on the similarity between the current and next states. At time steps where the state changes are small, and the similarity is high, a larger \(\Delta\) is chosen to accelerate the sampling process. Conversely, at time steps where the state changes are significant, and the similarity is low, a smaller \(\Delta\) is selected to capture details. The similarity is measured based on the absolute value of cosine similarity, calculated as follows:
\begin{equation}
\text{similarity} = \Bigg \lvert \frac{\hat{Y}_s \cdot \hat{Y}_{s-\Delta}}{\|\hat{Y}_s\| \|\hat{Y}_{s-\Delta}\|} \Bigg \rvert, \label{equ10}
\end{equation}
where \( s \in \{1, 2, \ldots, S\} \), \(\hat{Y}_s\) represents the predicted action sequence of the current state, and \(\hat{Y}_{s-\Delta}\) represents the predicted action sequence after skipping \(\Delta\) steps. Both sequences are flattened into one-dimensional vectors to calculate similarity. The symbol \(\cdot\) denotes the dot product of two vectors, and \(\|\hat{Y}_s\|\) and \(\|\hat{Y}_{s-\Delta}\|\) represent the Euclidean norms of vectors \(\hat{Y}_s\) and \(\hat{Y}_{s-\Delta}\), respectively. Based on the calculated similarity, the adjustment strategy for the step size \(\Delta\) can be described as:
\begin{equation}
\Delta_{\text{new}} = 
\begin{cases}
\Delta \times \gamma, & \text{if similarity} > \theta_{\text{high}} \\
\Delta, & \text{if } \theta_{\text{low}} \le \text{similarity} \le \theta_{\text{high}} \\
\Delta \div \gamma, & \text{if similarity} < \theta_{\text{low}}
\end{cases}, \label{equ11}
\end{equation}
where the adjustment factor \(\gamma > 1\), and \(\theta_{\text{high}}\) and \(\theta_{\text{low}}\) are preset upper and lower similarity thresholds, respectively, used to guide the increase, maintenance, or decrease of the step size. The adaptive skip strategy accelerates sampling by increasing the step size at time steps where the prediction results change minimally and optimizes prediction quality by decreasing the step size at time steps where the prediction results change significantly. This achieves an optimization of the inference process.

\begin{figure}[h]
    \centering
    \includegraphics[width=\textwidth]{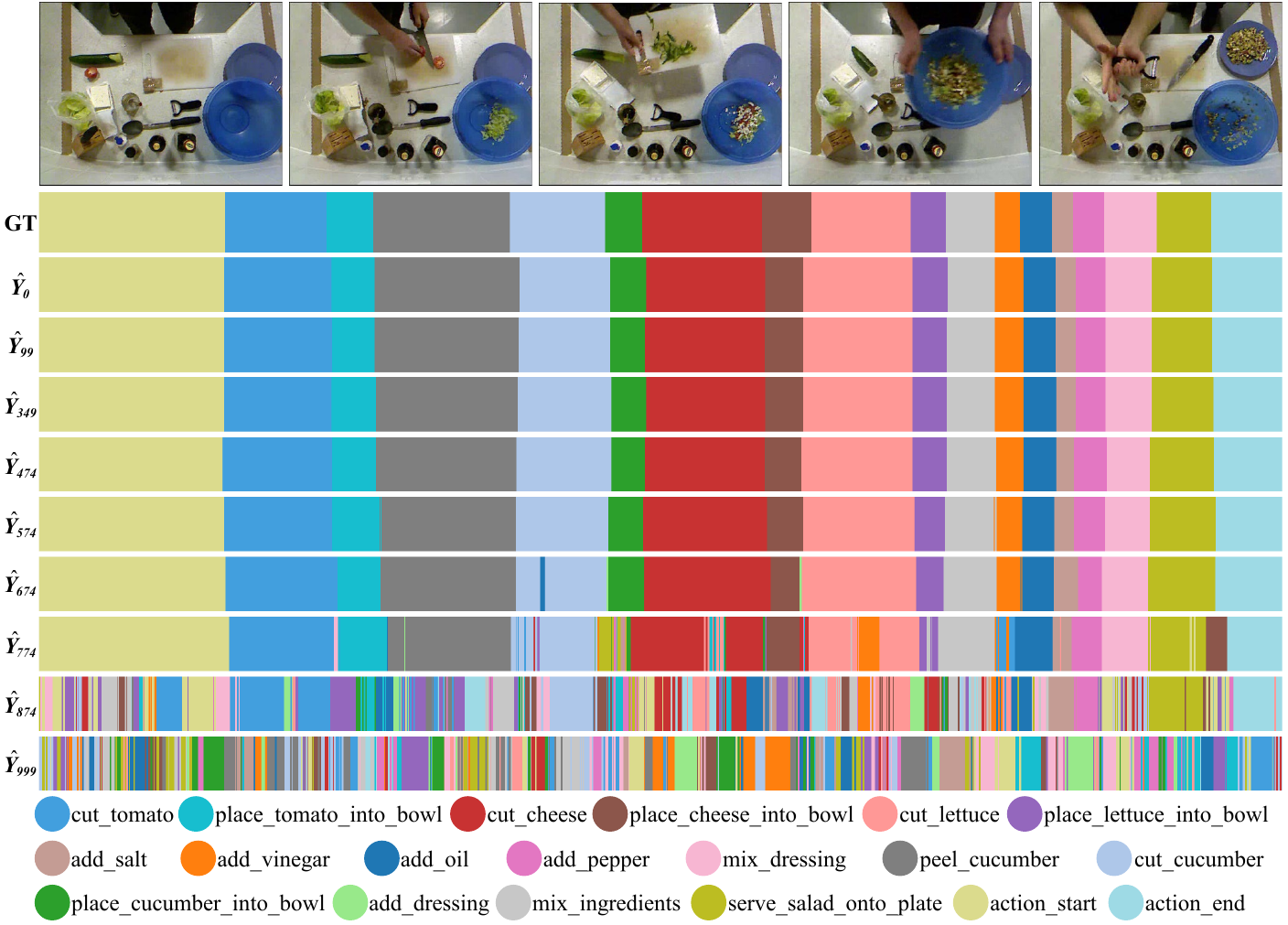}
    \caption{Example of the iterative denoising process for the adaptive skip strategy, where different colors in the diagram represent different action categories.}
    \label{fig6}
\end{figure}
\section{Experiments}

\subsection{Implementation Details}
For all datasets, this study uses the widely adopted I3D features \cite{carreira2017quo} as input features, with the feature dimension set to 2048. The encoder is constructed from the Temporal Dilated Perception (TDP) layers proposed in this paper, while the decoder follows the design from the literature \cite{liu2023diffusion}. The experimental framework is built on the PyTorch deep learning platform and executed in a single NVIDIA A800 GPU environment. To achieve end-to-end training, the Adam optimization algorithm is used, with a batch size of 4, and training is conducted for 5000 epochs. Specifically, for the Breakfast dataset, the learning rate is set to 1e-4, whereas for other datasets, the learning rate is set to 5e-4. The initial iteration count for model inference is set to 1000 steps, with an initial skip step length of 40 steps, which is adaptively adjusted based on the proposed adaptive skip strategy, as shown in Figure \ref{fig6}.

Data augmentation strategies are employed during the training and inference stages to enhance the model's generalization ability. Specifically, sampling starts from four different time points in the video sequence, and fixed-interval skip sampling is performed at each time point to generate four sub-sequences with slight temporal differences, resulting in four corresponding sub-prediction sequences. These sub-sequences are then recombined into a complete prediction sequence according to the original sampling order. Given the presence of a small amount of action noise in the prediction sequences, median filtering is applied to filter out the noise.

In experiments on the 50Salads \cite{50Salads}, Breakfast \cite{Breakfast}, and GTEA \cite{GTEA} public datasets, the Breakfast dataset is the largest, while 50Salads contains the longest videos and the most action segments in a single video. For the 50Salads dataset, a five-fold cross-validation method is used, while for the GTEA and Breakfast datasets, a four-fold cross-validation method is adopted, following the same data partitioning method as previous studies to ensure the comparability of experimental results.

\subsection{Datasets and Evaluation Metrics}
\textbf{50Salads Dataset.} The 50Salads dataset \cite{50Salads} includes 50 videos, each with an average length of six minutes. All were recorded from a top-down perspective, capturing the entire process of salad preparation in detail. A significant feature of this dataset is the diversity of participants and the fine-grained classification of action categories. The dataset captures 25 participants preparing two types of salads in the same kitchen environment. Each participant's actions are classified into 19 different action categories, providing a comprehensive perspective on this everyday task.

\textbf{Breakfast Dataset.}
The Breakfast dataset \cite{Breakfast} includes 1712 videos of breakfast preparation activities filmed from a third-person perspective. The dataset covers 18 different kitchen environments and provides detailed annotations for 48 action categories. These categories comprehensively cover all aspects of breakfast preparation. The average video length is approximately 2.3 minutes, with an average of 6.6 action instances per video.

\textbf{GTEA Dataset.}
The GTEA (Georgia Tech Egocentric Activity) dataset \cite{GTEA} is used for studying activity recognition and understanding from a first-person perspective. It records seven daily activities in detail, such as making a sandwich, tea, or coffee, and categorizes these activities into 11 action classes. Each activity was filmed by four different participants wearing a hat-mounted camera, resulting in 28 first-person-view videos. Each video is approximately one minute long, capturing around 20 fine-grained action segments, such as taking bread and pouring ketchup. The primary challenge of the GTEA dataset lies in its small data size and fine-grained action segments within each video.

\textbf{Metrics.}
Following previous work, the evaluation metrics include Frame-wise accuracy, Edit score, and F1 score. Frame-wise accuracy is a frame-level metric (action labels on a per-frame basis), while the Edit score and F1 score are segment-level metrics (aggregation of continuous actions).

\subsection{Comparisons with the State-of-the-art Methods}

Table \ref{tab1} presents the performance comparison of different methods for action segmentation tasks on three datasets. The performance of each method is evaluated using several key metrics, including F1 scores at different thresholds (10\%, 25\%, 50\%), Edit score, Frame-wise accuracy (Acc), and the average of these metrics (Avg).
\begin{table}[h]
\centering
\caption{Performance Comparison of EffiDiffAct and SOTA Methods on Several TAS Datasets.}
\label{tab1}
\vspace{0.2cm}
\resizebox{\linewidth}{!}{
\begin{tabular}{c|cccc|cccc|cccc}
\hline
 & \multicolumn{4}{c|}{50Salads} & \multicolumn{4}{c|}{Breakfast} & \multicolumn{4}{c}{GTEA} \\
\textbf{Model} & \textbf{F1@}\{\textbf{10, 25, 50}\} & \textbf{Edit} & \textbf{Acc} & \textbf{Avg} & \textbf{F1@}\{\textbf{10, 25, 50}\} & \textbf{Edit} & \textbf{Acc} & \textbf{Avg} & \textbf{F1@}\{\textbf{10, 25, 50}\} & \textbf{Edit} & \textbf{Acc} & \textbf{Avg}\\
\hline
\text{ASFormer}\cite{yi2021asformer} & 85.1 / 83.4 / 76.0 & 79.6 & 85.6 & 81.9 & 76.0 / 70.6 / 57.4 & 75.0 & 73.5 & 70.5 & 90.1 / 88.8 / 79.2 & 84.6 & 79.7 & 84.5 \\
\text{UARL}\cite{UARL} & 85.3 / 83.5 / 77.8 & 78.2 & 84.1 & 81.8 & 65.2 / 59.4 / 47.4 & 66.2 & 67.8 & 61.2 & 92.7 / 91.5 / 82.8 & 88.1 & 79.6 & 86.9\\
\text{DPRN}\cite{DPRN} & 87.8 / 86.3 / 79.4 & 82.0 & 87.2 & 84.5 & 75.6 / 70.5 / 57.6 & 75.1 & 71.7 & 70.1 & 92.9 / 92.0 / 82.9 & 90.9 & \textbf{82.0} & 88.1\\
\text{SEDT}\cite{SEDT} & 89.9 / 88.7 / 81.1 & 84.7 & 86.5 & 86.2 & - / - / - & - & - & - & \textbf{93.7} / \textbf{92.4} / 84.0 & 91.3 & 81.3 & \textbf{88.5}\\
\text{TCTr}\cite{TCTr} & 87.5 / 86.1 / 80.2 & 83.4 & 86.6 & 84.8 & 76.6 / 71.1 / 58.5 & 76.1 & 77.5 & 72.0 & 91.3 / 90.1 / 80.0 & 87.9 & 81.1 & 86.1\\
\text{FAMMSDTN}\cite{FAMMSDTN} & 86.2 / 84.4 / 77.9 & 79.9 & 86.4 & 83.0 & 78.5 / 72.9 / 60.2 & 77.5 & 74.8 & 72.8 & 91.6 / 90.9 / 80.9 & 88.3 & 80.7 & 86.5\\
\text{DTL}\cite{DTL} & 87.1 / 85.7 / 78.5 & 80.5 & 86.9 & 83.7 & 78.8 / 74.5 / 62.9 & 77.7 & 75.8 & 73.9 & - / - / - & - & - & -\\
\text{UVAST}\cite{UVAST} & 89.1 / 87.6 / 81.7 & 83.9 & 87.4 & 85.9 & 76.9 / 71.5 / 58.0 & 77.1 & 69.7 & 70.6 & 92.7 / 91.3 / 81.0 & \textbf{92.1} & 80.2 & 87.5\\
\text{LTContext}\cite{LTContext} & 89.4 / 87.7 / 82.0 & 83.2 & 87.7 & 86.0 & 77.6 / 72.6 / 60.1 & 77.0 & 74.2 & 72.3 & - / - / - & - & - & -\\
\text{DiffAct}\cite{liu2023diffusion} & 90.1 / 89.2 / 83.7 & 85.0 & 88.9 & 87.4 & 80.3 / 75.9 / 64.6 & 78.4 & 76.4 & 75.1 & 92.5 / 91.5 / 84.7 & 89.6 & 80.3 & 87.7\\
\text{EffiDiffAct (Ours)} & \textbf{91.7 / 91.0 / 87.1} & \textbf{87.1} & \textbf{90.2} & \textbf{89.4} & \textbf{81.7 / 77.3 / 66.5} & \textbf{79.7} & \textbf{78.3} & \textbf{76.7} & 92.6 / 91.8 / \textbf{85.3} & 89.7 & 81.1 & 88.1\\

\hline
\end{tabular}}
\end{table}

On the 50Salads dataset, the EffiDiffAct method demonstrates significant performance advantages. Specifically, EffiDiffAct achieves F1 scores of 91.7/91.0/87.1 at different thresholds (10\%, 25\%, 50\%), an Edit score of 87.1, and a Frame-wise accuracy (Acc) of 90.2. Combining all evaluation metrics, EffiDiffAct's average score is 89.4, significantly higher than DiffAct's score of 87.4. 
On this large-scale Breakfast dataset, EffiDiffAct also surpasses the SOTA methods in all key evaluation metrics, reflecting its generalization capability and effectiveness in fine-grained action recognition.
Considering the unique challenges of the GTEA dataset and the high segmentation accuracy achieved by existing methods, the performance improvement of the EffiDiffAct method on this dataset is insignificant. The experimental results shown in Table \ref{tab1} indicate that the EffiDiffAct method performs comparably to other studies and achieves the highest F1@50 score of 85.3, demonstrating its excellent capability in accurately identifying action start and end points. Overall, the performance across three datasets highlights the significant advantages of EffiDiffAct in several key metrics. On the more complex 50Salads dataset and the large-scale Breakfast dataset, EffiDiffAct exhibits more pronounced advantages, fully demonstrating its reliability and precision in action segmentation tasks.

\begin{figure}[h]
    \centering
    \includegraphics[width=1.0\textwidth]{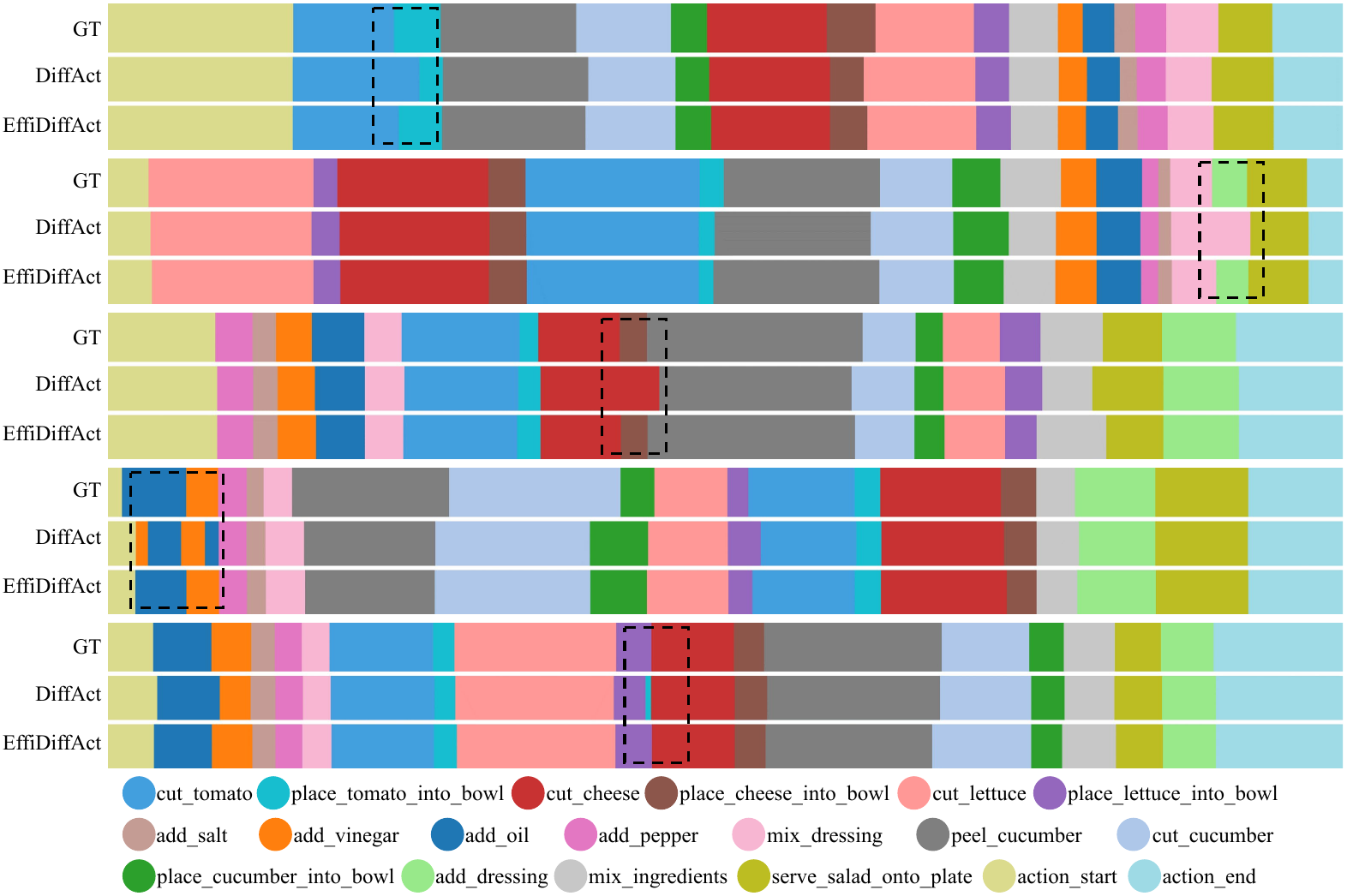}
    
    \caption{Qualitative analysis on the 50Salads dataset, where different colors represent different action categories.}
    \label{fig7}
\end{figure}

\subsection{Qualitative Analysis}
To more intuitively demonstrate the effectiveness of the proposed method, this study visualizes the comparison between the predicted results and the actual action labels (Ground Truth, GT) for five data samples, as shown in Figure \ref{fig7}. In the figure, different colors represent different action categories. In each sample, the top row shows the ground truth action labels, followed by the predicted results of the DiffAct and EffiDiffAct methods, respectively. The study then provides a detailed analysis of the visualization results based on three dimensions: boundary alignment, transient action recognition, and action segment continuity:

\begin{itemize}
\item Boundary Alignment: In the vast majority of samples, the predictions of EffiDiffAct exhibit closer alignment with the ground truth action labels (GT). In the circled area of Sample 1 (top sample), the action segments predicted by EffiDiffAct almost perfectly align with the GT, whereas DiffAct shows a significant deviation in the same area. This observation highlights the advantage of EffiDiffAct in accurately recognizing action boundaries.
   
\item Transient Action Recognition: In Samples 2 and 3, EffiDiffAct demonstrates high sensitivity to actions of shorter duration. Compared to the GT, EffiDiffAct more accurately identifies the start and end of these actions. In contrast, DiffAct exhibits issues with missing or under-segmenting these crucial transient actions.

\item Action Segment Continuity: Samples 4 and 5 show that EffiDiffAct maintains high consistency with the GT in capturing longer continuous actions. In comparison, DiffAct erroneously segments single continuous actions into multiple shorter segments.
\end{itemize}

\subsection{Ablation Studies}

To further validate the effectiveness of the proposed method, a series of ablation studies were conducted on the 50Salads dataset. This dataset was chosen because it has a high level of complexity and a moderate data size, making it suitable for an in-depth analysis of each component's contribution.

Firstly, the impact of different encoders on the model's performance was explored. As shown in Table \ref{tab2}, the performances of the Self-Attention, Windowed-Attention, LTContext-Attention, and TDP encoders were compared across several metrics. The results indicate that the TDP encoder outperforms other types of encoders in all evaluation metrics. Specifically, in terms of F1 scores, the TDP encoder achieved scores of 91.7/91.0/87.1 at 10\%, 25\%, and 50\% overlap thresholds, respectively, which are significantly higher than the highest scores of the Windowed-Attention encoder, 89.3/88.7/84.9, highlighting its advantage in action boundary recognition. Additionally, the TDP encoder's performance in the Edit and Frame-wise Accuracy (Acc) metrics further demonstrates its effectiveness in capturing complex action sequences.

\begin{table}[h]
\centering
\caption{Performance Comparison of EffiDiffAct Algorithm with Different Encoders.}
\label{tab2}
\vspace{0.2cm}

\begin{tabular}{c|cccc}
\hline
\text{Encoder} & \text{F1@}\{10, 25, 50\} & \text{Edit} & \text{Acc} & \text{Avg} \\
\hline \text{Self-Attention}\cite{vaswani2017attention} & 87.4/86.8/83.1 & 83.1 & 86.0 & 85.3 \\
\text{Windowed-Attention}\cite{LTContext} & 89.3/88.7/84.9 & 84.8 & 87.9 & 87.1 \\
\text{LTContext-Attention}\cite{LTContext} & 88.5/87.7/84.1 & 84.0 & 86.9 & 86.3 \\
\text{TDP} & \textbf{91.7}/\textbf{91.0}/\textbf{87.1} & \textbf{87.1} & \textbf{90.2} & \textbf{89.4} \\
\hline
\end{tabular}
\end{table}

Subsequently, a comparative analysis was performed between the proposed adaptive skip strategy and the fixed step strategy. As shown in Table \ref{tab3}, the adaptive skip strategy significantly reduces inference time while maintaining almost the same performance level compared to the fixed step strategy. With an initial iteration count of 25, the fixed step strategy skips a fixed time step length in each iteration. If the total time step length is 1000 and each iteration skips 40 steps, then after 25 iterations, the entire 1000-step range will be covered. In contrast, the adaptive skip strategy is more flexible, dynamically adjusting the step length of each iteration during inference. Thus, despite setting an initial iteration count of 25, the actual number of iterations might be less than 25. Thanks to the adaptive skip strategy, the inference time per video for EffiDiffAct on the 50Salads dataset decreased from 1.93 seconds to 1.65 seconds, a reduction of 14.5\%. This demonstrates that while the fixed step strategy is simple and straightforward, it lacks flexibility in practical applications, leading to wasted computational resources during inference. On the other hand, the adaptive skip strategy enhances computational efficiency while maintaining prediction quality by adaptively skipping unnecessary steps based on the similarity of consecutive states.

\begin{table}[h]
\centering
\caption{Performance and Inference Time Comparison Between Fixed Skip Strategy and Adaptive Skip Strategy.}
\label{tab3}
\vspace{0.2cm}

\resizebox{\textwidth}{!}{
\begin{tabular}{c|ccccc}
\hline
\text { Step Strategy } & \text { F1@\{10,25,50\} } & \text { Edit } & \text { Acc } & \text { Avg } & \text { Inference Time } \\
\hline \text { Fixed (25 Steps) } & \textbf{91.7 / 91.0 / 87.2} & \textbf{87.1} & \textbf{90.2} & \textbf{89.4} & 1.93 \textrm{s} \\
\text { Adaptive (25 Steps) } & \textbf{91.7 / 91.0 / 87.1} & \textbf{87.1} & \textbf{90.2} & \textbf{89.4} & \textbf{1.65} \textrm{s} \\
\hline
\end{tabular}}
\end{table}

\begin{table}[h]
\centering
\caption{Performance and Computational Cost Comparison Between EffiDiffAct and DiffAct Algorithms.}
\label{tab4}
\vspace{0.2cm}

\resizebox{\textwidth}{!}{
\begin{tabular}{c|cccc}
\hline
\text{Model} & \text{Avg} & \text{\#Parameters} & \text{Storage Size} & \text{Inference Time} \\
\hline 
\text{DiffAct (25 Steps)}\cite{liu2023diffusion} & 87.4 & 1.21 M & 4.81 MB & 2.10 \textrm{s} \\
\text{EffiDiffAct (8 Steps)} & 89.0 & \textbf{1.06} M & \textbf{4.23} MB & \textbf{0.76} \textrm{s} \\
\text{EffiDiffAct (16 Steps)} & 89.2 & \textbf{1.06} M & \textbf{4.23} MB & 1.18 \textrm{s} \\
\text{EffiDiffAct (25 Steps)} & \textbf{89.4} & \textbf{1.06} M & \textbf{4.23} MB & 1.65 \textrm{s} \\
\hline
\end{tabular}}
\end{table}

Additionally, this study compared the performance and computational cost of the EffiDiffAct method with the DiffAct method (see Table \ref{tab4}). Across different inference step settings, EffiDiffAct not only outperformed DiffAct in terms of performance but also demonstrated significant advantages in computational resource metrics such as the number of parameters, storage size, and inference time. Specifically, with an average performance of 87.4 for DiffAct, EffiDiffAct achieved an average performance of 89.0 with 8 inference steps, while reducing parameter count by 12.4\%, storage demand by approximately 12.1\%, and inference time by about 63.8\%. As the number of inference steps increased, EffiDiffAct's average performance further improved to 89.2 and 89.4 at 16 and 25 steps respectively. Although computational complexity and inference time also increased, EffiDiffAct still showed significant computational efficiency advantages compared to DiffAct. This analysis demonstrates that EffiDiffAct maintains high performance while improving overall computational efficiency.

\section{Conclusion}
This paper proposes EffiDiffAct, a more efficient and faster temporal action segmentation algorithm based on the diffusion framework, by designing a lightweight temporal feature encoder and an adaptive skip strategy. The lightweight encoder significantly reduces the computational burden of the model and effectively alleviates the issue of excessive smoothing of temporal features. The adaptive skip strategy optimizes the inference process by dynamically adjusting the step length, improving inference efficiency while maintaining prediction quality.  Extensive experiments on the 50Salads, Breakfast, and GTEA datasets, along with ablation experiments on the 50Salads dataset, confirm the effectiveness of the EffiDiffAct algorithm. The experimental results demonstrate that, compared to the SOTA methods, EffiDiffAct achieves significant improvements in both performance and computational efficiency in action segmentation tasks.

\section{Discussion}
Despite the outstanding performance of the EffiDiffAct algorithm in temporal action segmentation tasks, there are still some shortcomings.
Firstly, the algorithm showed significant improvements on large-scale datasets like 50Salads and Breakfast. However, its improvements on the small-scale GTEA dataset were limited. Future research should further address the generalization problem on small-scale datasets to ensure the robustness of the algorithm in various real-world application scenarios.
Secondly, EffiDiffAct's adaptive skip strategy relies on similarity metrics, and the computational complexity and accuracy of this process may be influenced by various factors. Future work could explore more efficient and accurate similarity measurement methods to further optimize the algorithm's performance.

\section{Acknowledgments}
This work was supported by the National Key R\&D Program of China
(2021ZD0113502); the Shanghai Municipal Science and Technology Major Project (2021SHZDZX0103).





\end{document}